# *Large language model and reinforcement learning based autonomous construction and progress updates*

1st Yifei Xiao
*Department of Civil Engineering*
*The University of British Columbia*
Vancouver, Canada
yifeix5@student.ubc.ca

2nd Huasheng Lin
*Department of Civil Engineering*
*The University of British Columbia*
Vancouver, Canada
huasheng@student.ubc.ca

3rd T.Y. Yang
*Department of Civil Engineering*
*The University of British Columbia*
Vancouver, Canada
yang@civil.ubc.ca

***Abstract*—Due to the rapid growth of the global population, traditional construction methods are gradually unable to meet the current housing demand. The advancement of machine learning algorithms has encouraged researchers to explore their applications in the construction industry, leading to more intelligent and efficient building processes. In particular, the rise of large language models in recent years has provided more possibilities for the transformation from traditional construction to smart construction. In this study, a large language model and reinforcement learning based autonomous construction method is proposed. An enhanced soft actor-critic reinforcement learning algorithm is developed to plan actions for a mobile crane according to construction tasks. Robot kinematics and control algorithms are used to ensure that the actions generated by the reinforcement learning algorithm can be executed by the crane. In addition, a customized embedding model is developed to extract information from the construction environment according to user interests. Then, a large language model, ChatGPT 5, is used to update construction progress, provide safety information to control system, and generate construction progress reports. The feasibility of proposed method is verified through several full-scall retaining wall construction experiments.**

***Keywords*—large language model, reinforcement learning, machine learning, robotic construction, smart construction.**

## I. INTRODUCTION

The traditional construction industry relied heavily on intensive on-site manual operations, and this construction method is facing increasingly severe challenges. The problem of labor shortage and outdated technology directly lead to lengthy project cycles, uncontrolled costs, and quality uncertainty [1-2]. These inherent points not only restrict the development of the construction industry, but also make it difficult to meet the urgent demand of modern society for fast and high-quality construction.

To overcome this dilemma, modular construction has emerged as an innovative approach. It has achieved a significant improvement in production efficiency and construction quality by transferring a large amount of construction activities from open and uncontrollable sites to enclosed and standardized factories. However, modularity itself is not the ultimate solution. Its development is still limited by the degree of automation in the prefabrication process within the factory and the level of intelligence in the on-site assembly process. To achieve the next leap in building production efficiency, it is necessary to promote a transformation from "industrialization" to "intelligence". The key to this transformation lies in the integration of machine leaning algorithms, mechanical automation and robotics technology [3]. At the factory end, robots can undertake component production and assembly tasks with high repeatability and strict precision requirements. At the construction site, robots can accurately lift, locate, and connect modules in complex and even dangerous environments, freeing workers from heavy physical labor. Through the integration of robotics and artificial intelligence, an autonomous construction system that integrates perception, decision-making, and execution can be established. Therefore, the development of robot assisted automated construction can fundamentally enhance the overall efficiency of the construction industry and provide key technical support for addressing future urbanization challenges [4].

## II. LITERATURE REVIEW AND OBJECTIVES

### *A. Large language model in construction*

The research towards application of large language model (LLM) in construction can be mainly divided into five categories: **(1) Project management:** LLM can perform necessary scheduling and resource allocation based on project task requirements and a large amount of project data. For example, LLM can utilize a large amount of historical data in a project to predict and provide task related plans. Meanwhile, LLM can also identify anomalies that have already occurred and take more proactive measures to avoid potential hazards in the future [5-6]. Therefore, adopting LLM predictive analysis method for risk management and decision-making can effectively reduce risks and increase construction efficiency. **(2) Design and building** **information modeling (BIM)**: BIM provides an platform to efficiently plan, design, and manage buildings and infrastructure projects. The application of LLM in BIM systems is mainly reflected in improving design efficiency, optimizing collaborative processes, and providing decision support. For example, LLM can be used to evaluate large amounts of data to understand what is most suitable for design, thereby improving decision-making speed and project success rate. Lee et al. introduced a generalized LLM-augmented BIM framework [7]. Through a step-by-step development process, LLM can enhance the development speed of BIM applications. **(3) Increasing productivity:** LLM can increase construction productivity across various domains, such as scheduling and resource management. LLM can assist in evaluating the layout of heavy machinery and equipment on site and allocating resources (such as manpower and materials) based on different construction tasks [8-9]. This will ensure a more streamlined workflow with less waste and increased productivity. **(4) Integrating Robotics for automation:** The combination of LLM and robotics can be used to perform repetitive tasks that require heavy physical labor. In robot control, LLM can serve as the brain of the robots, which undertake high-level intelligent decision-making and command parsing tasks. This enables robots to understand complex human commands, generate multi-step operation plans, and collaborate with low-level control systems to achieve more flexible, efficient, and autonomous robot behavior [10]. In automated construction, LLM can serve as the core of intelligent decision-making and project management. This helps to achieve autonomous planning, real-time monitoring, and collaborative scheduling of the construction process, thereby improving construction efficiency and safety [11]. **(5) Enhancing Safety:** LLM plays a crucial role in construction safety. Through intelligent analysis of sensor data, construction logs, and on-site reports, LLM can identify potential safety risks in real time and generate early warnings and optimization suggestions. In addition, LLM can assist in developing safety management strategies and automatically generating emergency plans, thereby enhancing the safety supervision level and ability to respond to emergencies on construction sites [12].

However, due to concerns about the uncertainties brought by technological transformation and the substantial financial investment required, the integration of artificial intelligence (AI) and LLM into the construction industry has remained relatively slow. The difficulty of integrating or aligning LLM with existing applications and machinery in the industry is a key challenge for building and construction companies to adopt this new technology. In addition, the maintenance and upgrade of AI and LLM systems also require an additional budget, making it increasingly difficult for small companies with limited resources to implement this cutting-edge technology and compete with large companies [13]. Furthermore, the construction industry has a certain dependence on skilled labor. However, with the introduction of AI and LLM technology, workers need to constantly improve their skills, especially mastering the latest technologies and data-driven methods. Finally, although researchers are attempting to use LLM to revolutionize the traditional construction industry, the vast majority of research was conducted and proved in simulation or laboratory environments.

### *B. Reinforcement learning in construction*

Reinforcement learning (RL), as a branch of machine learning, is able to train an agent to complete specific tasks through trial and error. In automated construction fields, RL are mainly involved in three aspects: **(1) Building design:** The application of RL in the field of building design is mainly reflected in the automatic generation of optimal planning solutions based on user-defined environments and constraints. For example, Liu et al. [14] employed Q-learning to achieve a more realistic path planning in steel rebar designs for real-world reinforced concrete structures. The feasibility of Q-learning algorithm was verified through analysis towards three typical beam-column joints and a two-story reinforced concrete building frame. Sahachaisaree, Sae-ma, and Nanakorn et al. [15] used actor-critic RL algorithm to optimize the design of a two-dimensional truss structure. The testing results indicated a significant improvement in computation time compared to traditional optimization methods. **(2) Construction management and maintenance:** RL in construction management can be used to optimize resource allocation, schedule arrangement, and equipment scheduling, thereby improving overall construction efficiency. It can also achieve more efficient equipment management and fault prevention by continuously learning on-site data and dynamically adjusting maintenance strategies. For example, Soleymani, Bonyani, and Attarzadeh [16] implemented double deep Q-networks to automatically allocate construction resources based on the resource information gathered from distributed Internet of Things (IoT) sensor devices. The results proved that the proposed method can efficiently allocate resources and be applied to large-scale resource management systems. Yao et al. [17] applied deep Q learning to learn a strategy that can improve the long-term cost-effectiveness of pavement maintenance. **(3) Robotic construction:** By implementing RL in robotic construction, robots can learn the optimal construction strategy through continuous interaction with the environment, achieving path planning, action control, and collaborative operations for complex tasks [18]. In addition, RL can also enhance the adaptability and autonomous decision-making ability of robots in dynamic construction environments, thereby promoting the development of intelligence and automation in the construction process [19]. For example, Huang and Zou [20] implemented proximal policy optimization (PPO) to train a robot arm and an unmanned ground vehicle to collaboratively transport and install window panel in sequence. Their proposed method was proved in a Pybullet simulation environment.

Although RL has shown great potential in robotic construction, there are still some limitations in its application. Firstly, the construction environment is complex and the ability of RL models in highly dynamic and unstructured scenarios is limited, making it difficult to directly transfer to actual construction sites. Secondly, the RL training process often requires a large amount of data and computing resources. There are safety and cost issues in obtaining high-quality samples in real construction environments. Furthermore, most existing studies towards application in RL still focus on simulation or small-scale experimental environments. The feasibility and practicality of RL in real construction projects still needs to be investigated.

### C. Objectives and scope

To address the current shortage of skilled labors in the construction industry and apply AI algorithms in real construction environments, a LLM and RL enabled automated construction framework is proposed in this study. In this framework, RL is used to plan lifting paths for a robotized crane to complete construction tasks. Robot kinematics and control algorithms are used to ensure that the crane can execute the lifting path. LLM is used to update construction progress, provide safety information to control system, and generate construction progress reports. To verify the proposed framework, a retaining wall construction experiment is carried out. The results of this study demonstrate the potential of using LLM and RL for real construction projects. It will also help the construction industry address the shortage of skilled labor and ensure higher quality control.

## III. Methodology

Fig. 1 shows the LLM and RL enabled automated construction framework. Based on the previous achievements of our research team [21], an enhanced soft actor-critic (SAC-E) RL algorithm is used to plan a lifting path for the robotized crane to transport construction materials to target positions. This lifting path is composed of a series of three-dimensional (3D) coordinates and each 3D coordinate is analyzed through robot kinematics and control algorithms to generate control signals to precisely control crane motions. In this study, the proportional-integral-derivative (PID) is implemented as the controller to control the robotized crane. After the crane is controlled to reach the target position, the crane states are updated and the updated crane states are used as the inputs for the SAC-E RL algorithm. On the other hand, LLM receives the results from robot kinematics, the environmental information obtained from original environment image, and the embedded environmental information to update construction progress, provide safety information to control system, and generate construction progress reports. At the same time, safety information generated by the LLM will be sent to the control system to determine whether to suspend the construction task.

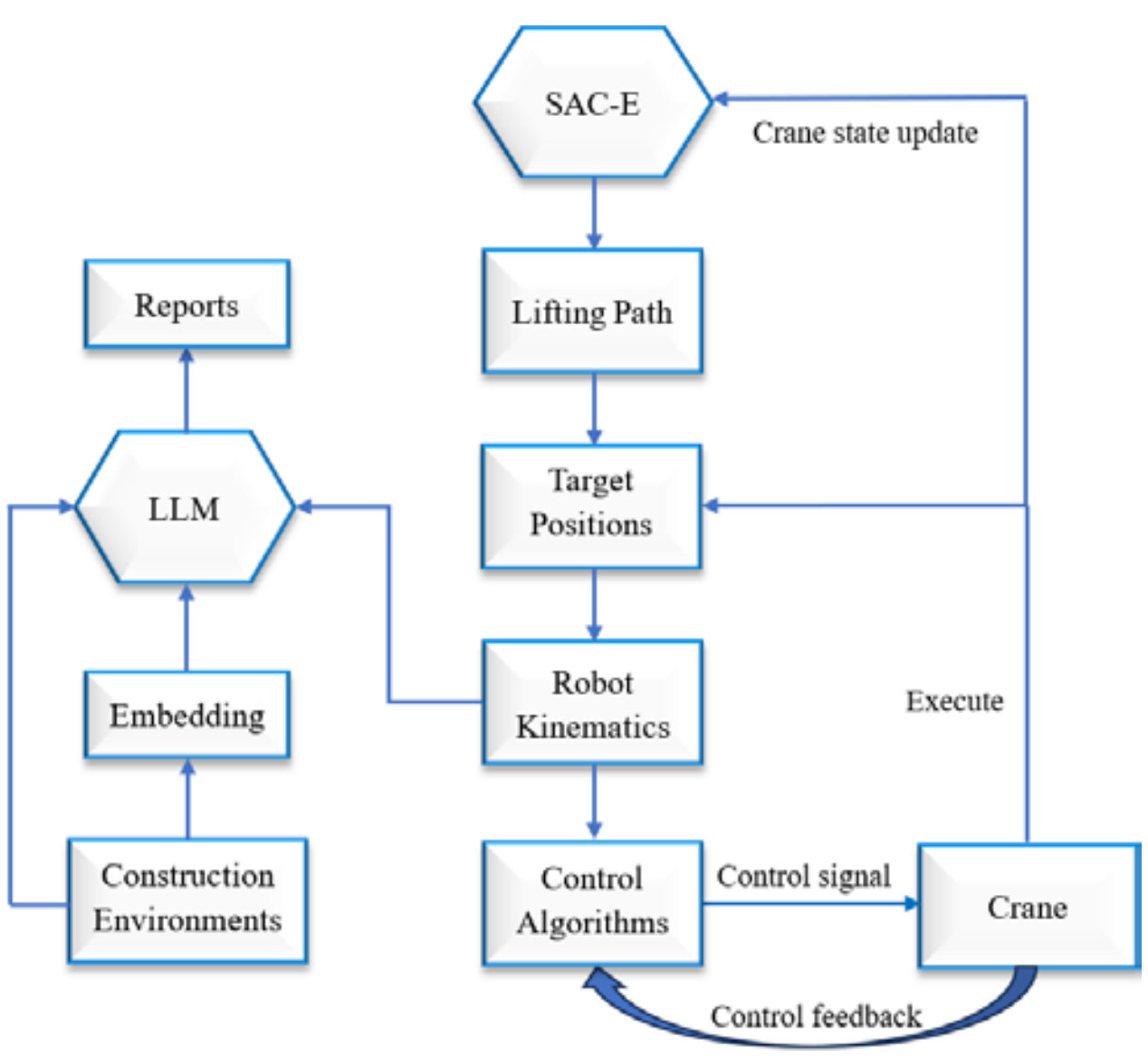

Fig. 1. LLM and RL enabled automated construction framework

### A. SAC-E RL algorithm

Traditional soft actor-critic (SAC) RL algorithm utilizes entropy regularization and softening strategy to improve learning efficiency and encourage agents to better explore unknown states [22]. In SAC, three neural networks are deployed, including a state value Q-network, a target Q-network, and an actor network. The actor network is used to generate actions for the agent to execute. The state value Q-network provides evaluations for the outputs from actor network. The target Q-network is used to provide a stable target value to train the state value Q-network, ensuring a more stable training process. However, in this research project, the traditional SAC RL algorithm is not as stable as desired during the training process, and it also has a high probability of failure when applied in reality. To improve stability and scalability of traditional SAC RL algorithm, a number of approaches were adopted by our research team to enhance the performance, including adaptive temperature, random sampling, learning rate decay, and reward scaling [21]. By implementing the approaches above, our research team has proved that SAC-E can achieve a more stable training process and a stronger scalability. The detailed SAC-E training process can be found in our previous study [21], and the well-trained model will be implemented in the experiment presented in this study.

### B. Customized embedding model for LLM analysis

The Segment Anything Model (SAM) is a category-agnostic segmentation framework capable of delineating objects based on point or bounding-box prompts. Instance masks and stable identifiers are first extracted for key on-site entities, including the moving block, workers, obstacles, and the target position. Based on these masks, geometric features are computed, and each frame is serialized into a compact state vector ($p_n$, $N_{ob}$, $S_{ob}$, $S_{warn}$), which represents task progress, the number of obstacles, the obstacle-crossing stage, and safety warning status, respectively. This structured representation enables real-time tracking and safety assessment. Then, LLM performs further analysis on this structured vector and combines the original environment image with the results of robot kinematics to obtain data that better meets our expectations. This combination can make the analysis results from LLM more focused on the data we need, rather than all the information (includes irrelevant information) contained in the entire construction environment.

By employing the Segment Anything approach, multiple objects of interest within the construction environment can be simultaneously localized and tracked, including the moving block, workers, and the destination point. The work progress percentage ($p_n$) is defined according to the horizontal distance between the current position of the transported object and its designated target position.

$$p_n = 1 - \left(\frac{x_{t,n} - x_{des}}{x_{t,0} - x_{des}}\right) \tag{1}$$

where $x_{t,0}$ is the original position of the transported object. $x_{t,n}$ is the real-time position of the transported object during lifting process. $x_{des}$ is the target position that the transported object needs to be placed.

The number of obstacles ($N_{ob}$) represents how many distinct obstacles are detected in the current scene. The obstacle-crossing stage, $S_{ob}$, takes values in {0,1,2} based on the status of obstacle avoidance: $S_{ob}$=0 (not crossing), $S_{ob}$=1(crossing), and $S_{ob}$=2 (crossing completed). To ensure safety, a safety warning index ($S_{warn}$) is defined. The safety warning index is triggered if workers are detected in images. This framework allows real-time monitoring of work progress and safety status through visual tracking and geometric relationships between segmented targets. By using the vision model, each environment monitoring picture is embedded into the environment vector: $(p_n,\ N_{ob}, S_{ob}, S_{warn})$.

### C. Construction progress update using LLM

Fig. 2 plots a pipeline of construction progress update using LLM. The prompt of LLM are composed of four parts, including a task description, results from robot kinematics, original construction environment, and the embedded environment vector $\boldsymbol{V_{ec}} = (p_n, N_{ob}, S_{ob}, S_{warn})$ with examples provided. The benefits of using the original construction environment image and the customized environment vector $\boldsymbol{V_{ec}}$ in LLM include: (1) The customized embedded model allows LLM to extract specific information and analyze it in a targeted manner. (2) The original environment image can give LLM a basic understanding of the overall construction environment (provides the "context of the environment"). (3) Customized embeddings can be reused, indexed, and retrievable. (4) The final analysis result from LLM is more in line with the user's intention. The role of the LLM, the specific task requirement, and the format of the generated data are defined in task description. After the LLM analyzes the results based on the prompt, the construction progress can be updated and the construction reports can be generated. Safety information generated by the LLM will be sent to the control system, and the control system will determine whether to suspend the task.

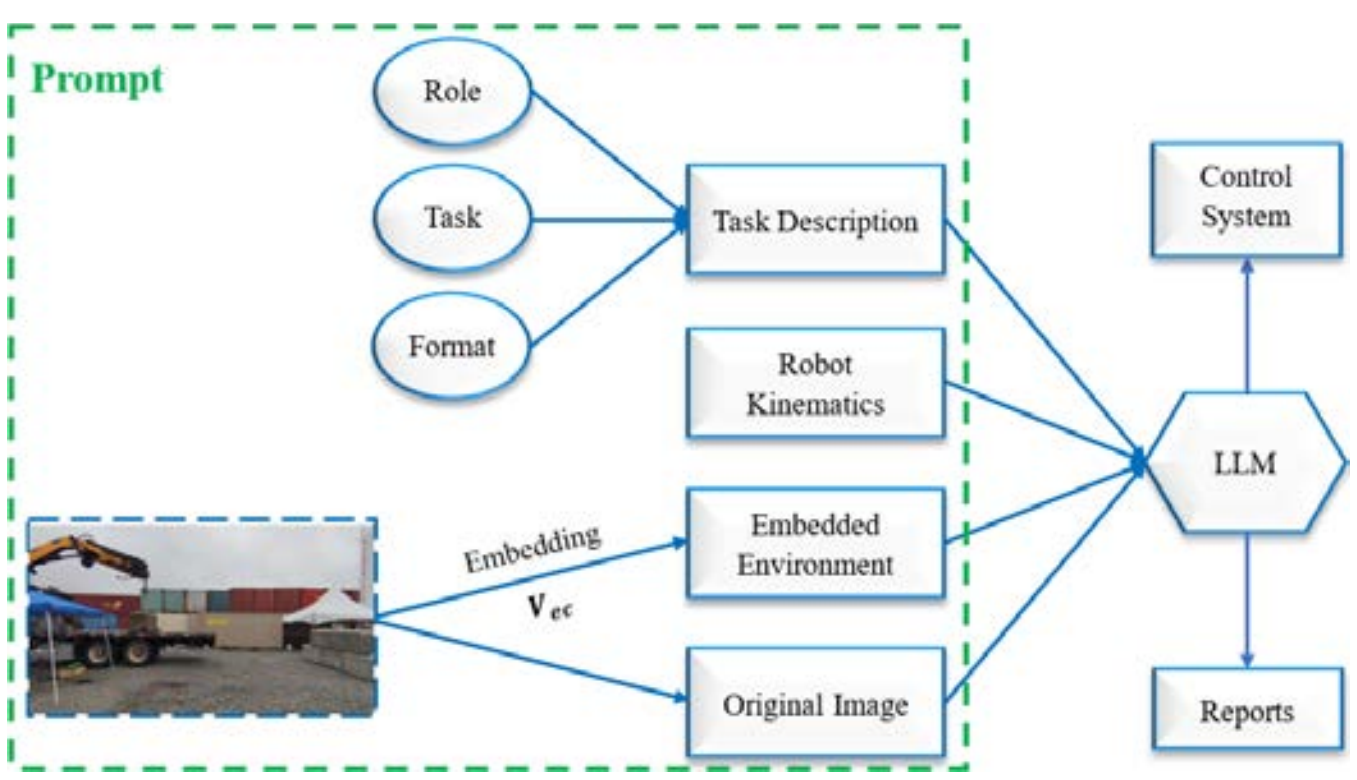

Fig. 2. Pipeline of construction progress updates using LLM

## IV. Experiment

### A. Experiment setup

To implement the proposed LLM and RL enabled automated construction framework, a high-performance computer equipped with Intel Ultra 9 285K and RTX 5090 is used to run the LLM model, SAC-E RL model, and robot kinematics algorithm, as shown in Fig. 3. An advanced control system (ACS) is used to run the robotic kinematics algorithm and the PID controller. A data acquisition (DAQ) system is used to send control signals to the robotized crane and receive the sensor feedback for the PID controller. To verify the feasibility of the proposed LLM and RL enabled automated construction framework, a COPMA 430_ENG mobile crane which was robotized by our research team was used to undertake a full-scale concrete retaining wall construction experiment. The task for the robotized crane is to transport a concrete block from the crane flatbed to the retaining wall structure with obstacle avoidance. The size of the concrete block is 1.6 × 0.8 × 0.8 m and the weight is 2400 kg. String potentiometers were installed on the crane turntable, inner boom and outer boom to measure the crane joint rotation. A laser sensor was installed on the telescopic to measure the extension of the robotized crane.

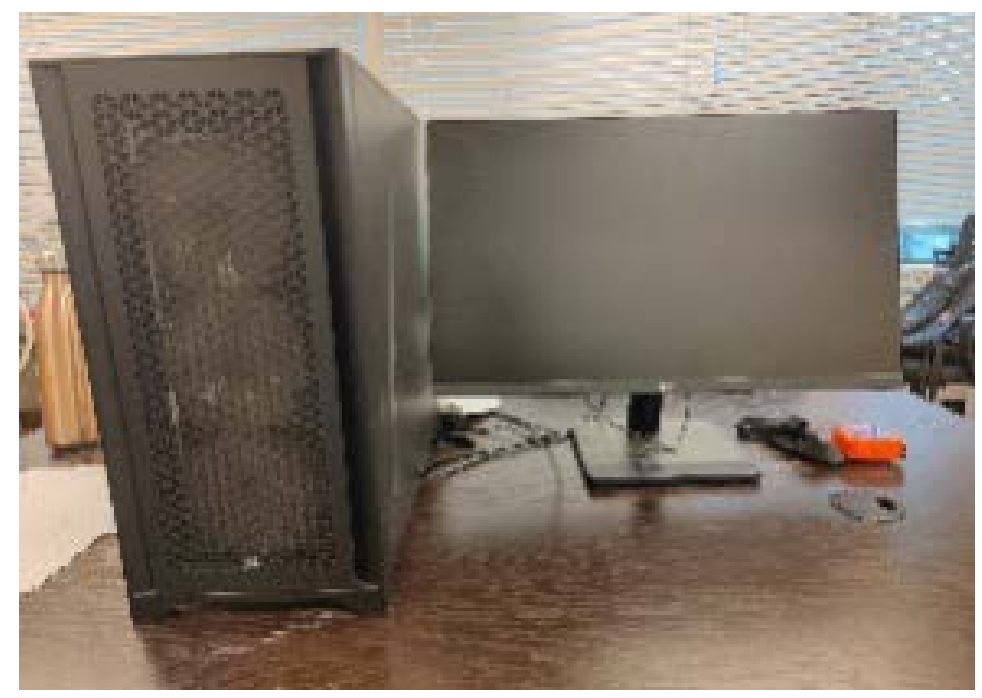

(a) High-performance computer

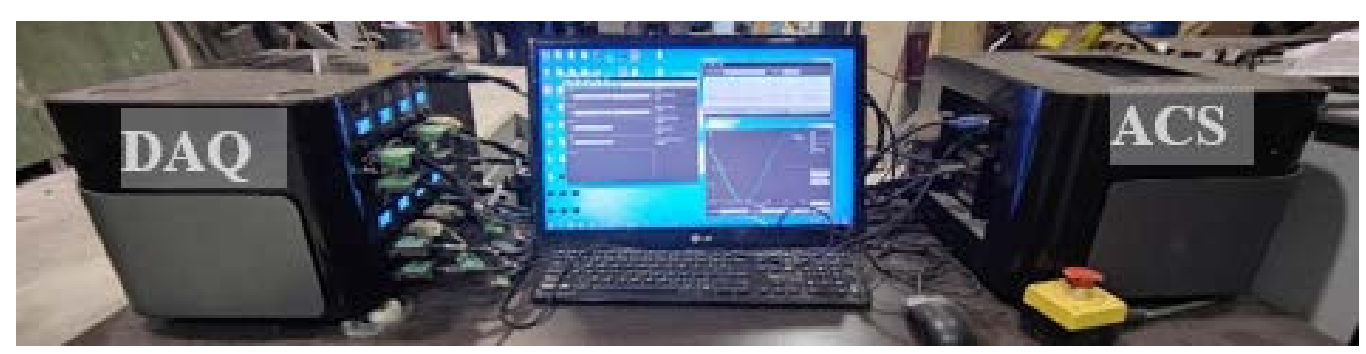

(b) High-performance computer

Fig. 3. Computational devices for the experiment

### B. Construction of retaining wall

To demonstrate the feasibility of the proposed LLM and RL enabled automated construction framework, several retaining wall construction experiments were conducted. The robotized COPMA 430_ENG mobile crane was used to undertake the concrete block transportation tasks. During the experiment, the SAC-E RL model was first used to plan the lifting path for the robotized crane to execute. Each 3D coordinate on the lifting path was analyzed by robot kinematics (inverse kinematics) to generate reference signals (required joint movements for all joints of the robotized crane) for PID control algorithm. Then, the PID algorithm can ensure that that each joint of the robotized crane can move to its reference position. When all joints move to the reference positions, the concrete block was transported to the current analyzed 3D coordinate. After all 3D coordinates were executed, the entire lifting path planned by SAC-E was executed. During experiment, it took approximately 4 minutes for each transporting task. Due to the implementation of PID control (a linear control algorithm), there were approximately 6 failures out of 35 experiments, resulting in a success rate of about 83%. In these failed cases, the concrete blocks were not fully placed in the target position due to small errors (approximately 1-3cm). However, such errors are generally permissible during crane hoisting process and can be eliminated by human intervention, such as by having someone assist in positioning the blocks at the target location.

In this study, ChatGPT 5 was used as the LLM to generate construction reports. Due to the slow operation speed of the robotized crane compared to computational speed of SAC-E and ChatGPT 5, ChatGPT 5 is able to generate a simple progress analysis report after one 3D coordinate is executed by the crane and provide feedback on the safety information to control system. When receiving the embedded environment vector $\boldsymbol{V_{ec}}$, ChatGPT 5 can determine whether $S_{warn}$ needs to be changed based on the position of the person in the image. To be specific, if the customized embedding model outputs $S_{warn} = 1$, it indicates that a person has been detected in the image. However, if this person is outside the construction zone or is sufficiently far from the crane, this situation should still be classified as safe. In such cases, ChatGPT 5 can assess the person's location in the image to determine whether $S_{warn}$ should indeed be set to 1. If ChatGPT 5 ultimately concludes that $S_{warn} = 1$, construction will be suspended and will only resume after the person has left the area. Fig. 4 and Fig.5 present the process of retaining wall construction. As can be seen from Fig.5, people were detected in both the first and fourth images. However, $S_{warn}$ is defined as 1 only when people were detected in the construction area (first image). During experiment, the latency between the appearance of people within the construction area to the suspension of crane operation is approximately 2-3 seconds. It was observed that the lifting process was suspended during the concrete block lifting stage because people were detected within the construction area ($S_{warn} = 1$). The construction process continued after no one was detected within the construction area. Fig. 6 shows an example of the prompt used in this study, which includes a task description, results from robot kinematics, original construction environment, and the embedded environment vector $\boldsymbol{V_{ec}} = (p_n, N_{ob}, S_{ob}, S_{warn})$. Experimental results have shown that the proposed LLM and RL enabled automated construction framework can achieve autonomous construction with real-time progress updates in both obstacle-free and obstacle-present scenarios.

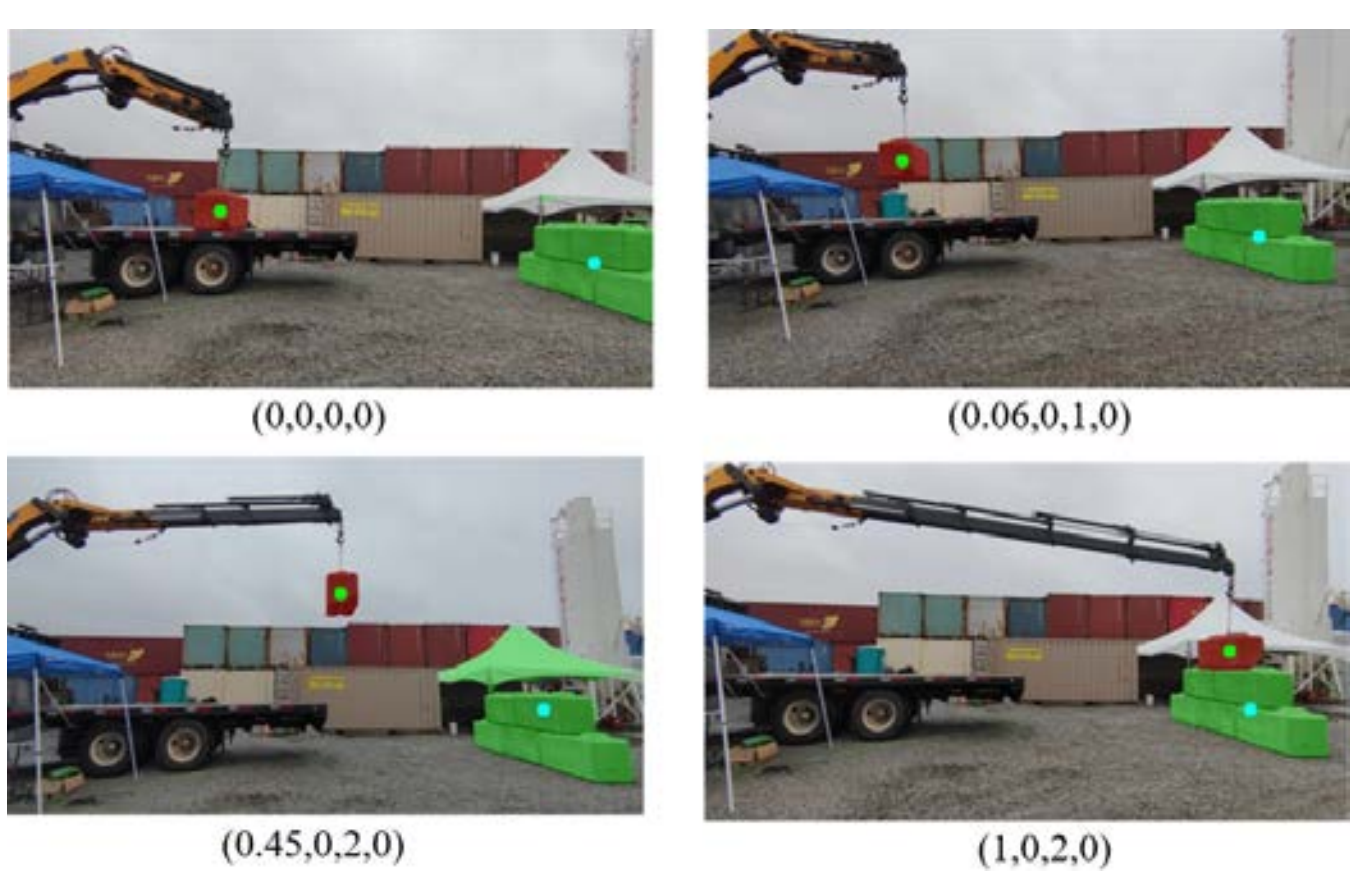

Fig. 4. Rretaining wall construction without obstacle

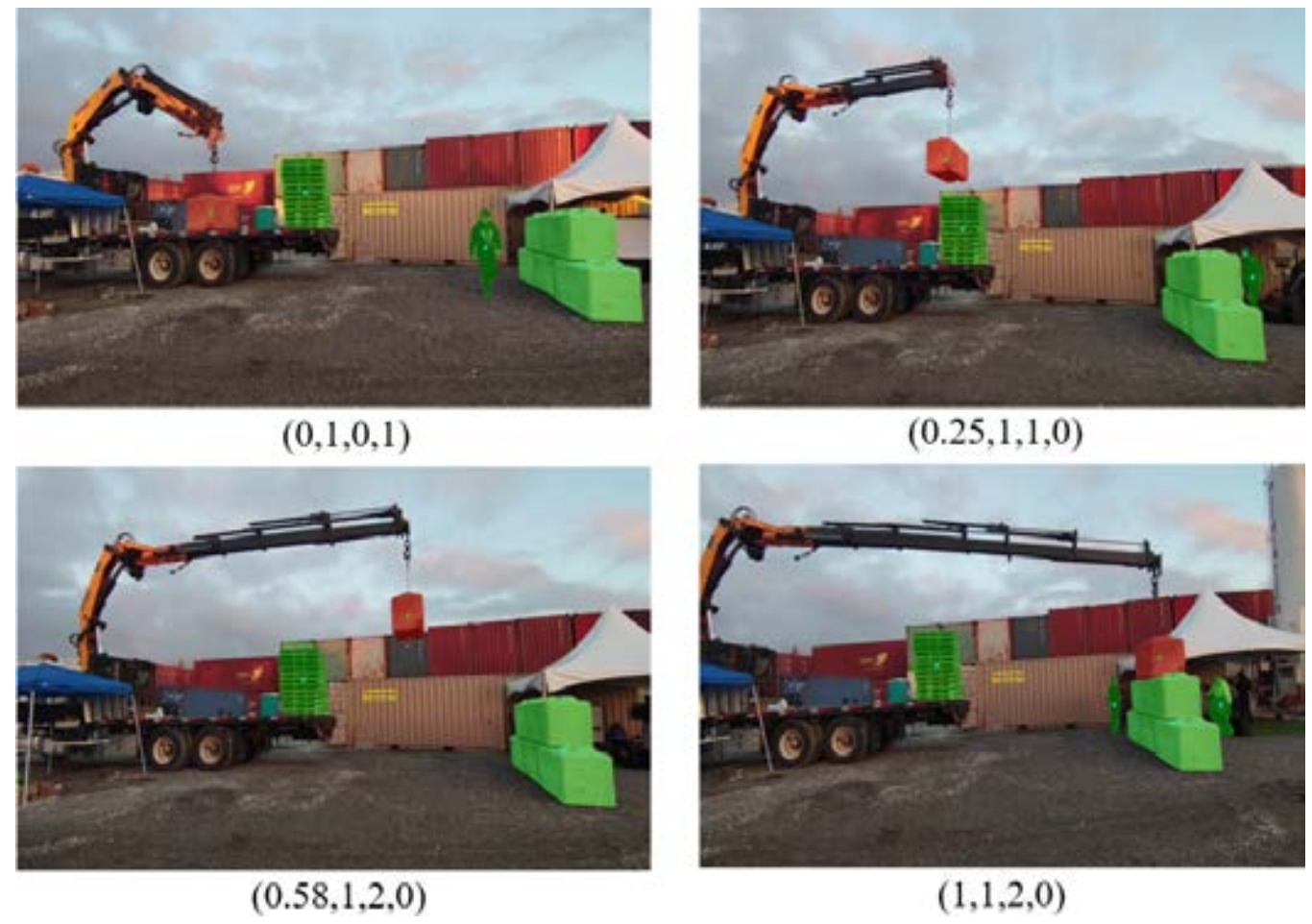

Fig. 5. Retaining wall construction with obstacle

```
messages = [
    {
        "role": "system",
        "content": (
            "You are a Construction Progress Analysis System employed by a project management team."
            "Analyze visual data and infer progress accurately."
        ),
    },
    {
        "role": "user",
        "content": [
            {
                "type": "image",
                "image": "D:/LLM_qwen2_5/construction/site1.jpg",
            },
            {
                "type": "text",
                "text": (
                    "A concrete block is being transported to the third level of a retaining wall. "
                    "Based on the image, estimate what percentage of the transportation task has been completed,"
                    "and provide reasoning with supporting details."
                    "Provide estimation based on following information"
                    "and the detailed explanation provided in extra_prompt section:"
                    "1) Uploaded on-site construction photos"
                    "2) Customized embedded environment vector s_n: [p_n, N_ob, S_ob, S_warn]"
                    "3)  Results from robot kinematics, including all joint movements of a mobile crane"
                ),
            },
```

```
extra_prompt = """Symbols (fixed) of s_n:
- p_n: work progress percentage
- N_ob: number of obstacles detected in the current frame.
- S_ob ∈ {0,1,2}: obstacle-crossing stage
   - 0 = not crossing (lifting)
   - 1 = crossing
   - 2 = crossing completed
- S_warn ∈ {0,1}: safety warning
   - 0 = safe
   - 1 = Unsafe (people are detected in the construction area or too closed to the machinery)

Criteria for construction progress classification (apply in order):
1) Safety:
   - If S_warn = 1 → safety_action = SUSPEND/HUMAN-TAKEOVER
   - Else → safety_action = CONTINUE
2) Stage selection:
   - If S_ob = 1 → stage = OBSTACLE-AVOIDANCE
   - Else if p_n is near completion (e.g., ≥ 95%) and S_ob = 2 → stage = PLACING
   - Else if S_ob = 2 → stage = FREE-TRANSPORT
   - Else (S_ob = 0) → stage = LIFTING

Output (strict JSON):
{
  "stage": "LIFTING | OBSTACLE-AVOIDANCE | FREE-TRANSPORT | PLACING",
  "safety_action": "CONTINUE | STOP/HUMAN-TAKEOVER",
  "num_obstacles": "INTEGER (number of detected obstacles in the scene)"
  "Current_progress": "percentage"
}
```

Fig. 6. Example of the prompts in this study

### C. *Ethical Considerations*

The deployment of reinforcement-learning-based and LLM-assisted robotic systems in construction raises important ethical issues, particularly because civil engineering projects involve high-risk environments and direct impacts on workers and the public. For this purpose, the proposed system has established multiple safeguard mechanisms to ensure safe and reliable operations: (1) The crane operation is still monitored by professional crane operators to ensure that automated operations can be intervened by humans at any time, and key decision-making remains attributable to human operators. (2) All sensing and data collection comply with on-site privacy protocols. No personal data are collected, stored, or processed except for information necessary for device perception and control. (3) The automation is designed to enhance worker safety by reducing exposure to hazardous tasks, rather than replacing skilled labors, aligning with ongoing industry practices that emphasize the importance of workers.

## V. CONCLUSION

Nowadays, traditional construction industry is facing severe challenges, such as the shortage of skilled workers and a serious lag in technology innovation. To address the labor shortage issues and bring new technology to the construction industry, a LLM and RL enabled automated construction framework is proposed in this study. In this framework, SAC-E RL algorithm is deployed to plan a lifting path for a robotized crane to complete construction tasks. Robot kinematics and control algorithm are used to ensure that the crane can execute the lifting path. In addition, a customized embedding model is developed to provide precise indicators (an embedded environment vector) according to user interests. Then, ChatGPT 5 is implemented to update construction progress, provide safety information to control system, and generate construction progress reports based on the results from robot kinematics, the original environment image, and the embedded environment vector. The proposed framework is verified through several full-scale retaining wall construction experiments, and the results have shown that the proposed framework can achieve autonomous construction with real-time progress updates in both obstacle-free and obstacle-present scenarios.